\pdfoutput=1

\documentclass[11pt]{article}

\usepackage{emnlp2021}

\usepackage{times}
\usepackage{latexsym}
\usepackage{graphicx} 

\usepackage{amsmath}
\usepackage{cleveref}
\crefformat{section}{\S#2#1#3} 
\crefformat{subsection}{\S#2#1#3}
\crefformat{subsubsection}{\S#2#1#3}

\usepackage[T1]{fontenc}

\usepackage[utf8]{inputenc}

\usepackage{microtype}

\title{Contrastive Learning for Context-aware Neural Machine Translation Using Coreference Information}

\author{Yongkeun Hwang\footnotemark[1], Hyungu Yun\footnotemark[1], Kyomin Jung\footnotemark[1] \footnotemark[2] \\
  \footnotemark[1] Dept. of Electrical and Computer Engineering, Seoul National University, Seoul, Korea \\
  \footnotemark[2] Automation and Systems Research Institute, Seoul National University, Seoul, Korea \\
  \{wangcho2k, youaredead, kjung\}@snu.ac.kr
}
\begin{document}
\maketitle
\begin{abstract}
Context-aware neural machine translation (NMT) incorporates contextual information of surrounding texts, that can improve the translation quality of document-level machine translation. Many existing works on context-aware NMT have focused on developing new model architectures for incorporating additional contexts and have shown some promising results. However, most existing works rely on cross-entropy loss, resulting in limited use of contextual information. In this paper, we propose CorefCL, a novel data augmentation and contrastive learning scheme based on coreference between the source and contextual sentences. By corrupting automatically detected coreference mentions in the contextual sentence, CorefCL can train the model to be sensitive to coreference inconsistency. We experimented with our method on common context-aware NMT models and two document-level translation tasks. In the experiments, our method consistently improved BLEU of compared models on English-German and English-Korean tasks. We also show that our method significantly improves coreference resolution in the English-German contrastive test suite.
\end{abstract}

\section{Introduction}
Neural machine translation (NMT) has achieved impressive performances on translation quality, due to the introduction of novel deep neural network (DNN) architectures such as encoder-decoder model \cite{cho-etal-2014-learning,Sutskever2014Sequence}, and self-attentional networks like Transformer \cite{Vaswani2017Attention}. The state-of-the-art NMT systems are now even comparable with human translators in sentence-level performance. 

However, there are a number of issues on document-level translation \cite{laubli-etal-2018-machine}. These include pronoun resolution across the sentences \cite{guillou-etal-2018-pronoun}, which needs cross-sentential contexts. To incorporate such document-level contextual information, several methods for context-aware NMT have been recently proposed. Many of the works have focused on introducing new model architectures like multi-encoder models \cite{voita-etal-2018-context} for encompassing contextual texts of the source language. These works have shown significant improvement in addressing discourse phenomena such as anaphora resolution mentioned above, as well as moderate improvements in overall translation quality.

Despite some promising results, most of the existing works have trained the model by minimizing cross-entropy loss, making the model rather exploit contextual information implicitly such as a form of regularization \cite{kim-etal-2019-document, li-etal-2020-multi-encoder}. Data augmentation for context-aware NMT is also an important issue, despite that recent works have focused on back-translation \cite{huo-etal-2020-diving}.

In this paper, we propose a Coreference-based Contrastive Learning for context-aware NMT (CorefCL), a novel data augmentation and contrastive learning scheme leveraging coreference information. Cross-sentential coreference between the source and target sentence can be a good source of training signal for context-aware NMT since it occurs when one or more expressions refer to the same entity, thus reflects dependencies between the source and contextual sentences.

CorefCL starts by conducting automatic annotation of coreference between the source and contextual sentences. Then, the referred mentions on contextual sentences are corrupted by removing and/or replacing tokens to generate contrastive examples. 
With those contrastive examples, we introduce a contrastive learning scheme equipped with a max-margin loss which encourages the model to discriminate between the original examples and the contrastive ones.
By doing so, CorefCL makes the model more sensitive to cross-sentential contextual information.

We experimented with CorefCL on three English-German corpora and one English-Korean document-level corpus, including WMT, IWSLT TED talk, and OpenSubtitles'18 English-German subtitles translation, and a web-crawled English-Korean subtitles translation. In all translation tasks, CorefCL consistently improves overall BLEU over baseline models without CorefCL. On experiments with three common context-aware model settings, we show that improvements by CorefCL are also model-agnostic. Finally, we show that the proposed method significantly improved the performance on ContraPro \cite{muller-etal-2018-large}, an English-German contrastive coreference benchmark.

\section{Related Works}

\subsection{Context-aware NMT}
Context-aware machine translation has been vigorously studied to exploit the crucial context information in surrounding sentences.
Recent works have shown that contextual information can help the model to generate not only more consistent but also more accurate translation \cite{smith2017integrating,voita-etal-2018-context,muller-etal-2018-large,kim-etal-2019-document}. 

In particular, \citet{voita-etal-2018-context} introduced a context-aware Transformer model which is able to induce anaphora relations, \citet{miculicich-etal-2018-document} showed that a model using cross-sentential contextual information significantly outperforms in document-level translation tasks, and \citet{Yun-Hwang2020Improving} insisted that context-aware models record the best performance especially in spoken language translation tasks where mandatory information tend to be sparse over multiple sentences.

The simplest method for context-aware machine translation is to concatenate all surrounding sentences and treat the concatenated sequence as a single sentence \cite{tiedemann-scherrer-2017-neural}. 
Although the concatenation strategy boosted Transformer architectures in multiple tasks \cite{tiedemann-scherrer-2017-neural,voita-etal-2018-context,Yun-Hwang2020Improving}, it lagged behind efficiency as the Transformer architecture has limited long-range dependency \cite{tang-etal-2018-self}.

To improve the efficiency, an additional encoder module is introduced to encode only the context sentences \cite{voita-etal-2018-context,jean2017does}. 
Additionally, hierarchical structures also have been introduced because the context sentences do not have the same significance as the input sentences \cite{miculicich-etal-2018-document,Yun-Hwang2020Improving}.

\subsection{Coreference and NMT}
The difference in coreference expressions among languages \cite{zinsmeister2017abstract, lapshinova-koltunski-etal-2020-coreference} gives MT systems a challenge on pronoun translation. Several recent works have attempted to incorporate coreference information \cite{ohtani-etal-2019-context}. The closest work to ours is \cite{stojanovski-fraser-2018-coreference} which also adds noises on creating a coreference-augmented dataset, while we do not add oracle coreference information directly to the training data.

\subsection{Data augmentation for NMT}
One of the most common methods for data augmentation in NMT is back-translation that generates pseudo-parallel data from monolingual corpora using intermediate NMT models \cite{sennrich-etal-2016-improving}. Generally, back-translation is conducted at sentence-level, however, several works have proposed document-level back-translation \cite{sugiyama-yoshinaga-2019-data, huo-etal-2020-diving}. 

On the other hand, sentence corruption by removing or replacing word(s) has also been widely used for improving model performance and robustness \cite{lample-etal-2018-phrase, voita-etal-2019-good}. Inspired by these works, we choose sentence corruption for contrastive learning.

\subsection{Contrastive Learning}
Contrastive learning is to learn a representation by contrasting positive and negative (contrastive) examples. It has been succeed in various machine learning fields including computer vision \cite{chen2020simple} and natural language processing \cite{mikolov2013distributed, wu2020clear, Lee2021contrastive}. 

Recently, several approaches on contrastive learning for NMT have also been studied. \citet{yang-etal-2019-reducing} proposed strategies for generating word-omitted contrastive examples and leveraging contrastive learning for reducing word omission errors on NMT. \citet{pan-etal-2021-contrastive} applied contrastive learning for multilingual MT and employed data augmentation for obtaining both the positive and negative training examples. 

While these works have been conducted on sentence-level NMT settings, we focus on extending contrastive learning on context-aware NMT. 

\section{Context-aware NMT models}
In this section, we briefly overview context-aware NMT methods and describe our baseline models which are also commonly adopted in recent works. 

Generally, a sentence-level (context-agnostic) NMT model takes an input sentence in a source language and returns an output sentence in a target language. On the other hand, a context-aware NMT model is designed to handle surrounding contextual sentences of source and/or target sentences. We focus on leveraging the contextual sentences of the source language.

\begin{figure}[ht]
    \resizebox{0.95\columnwidth}{!}{\includegraphics{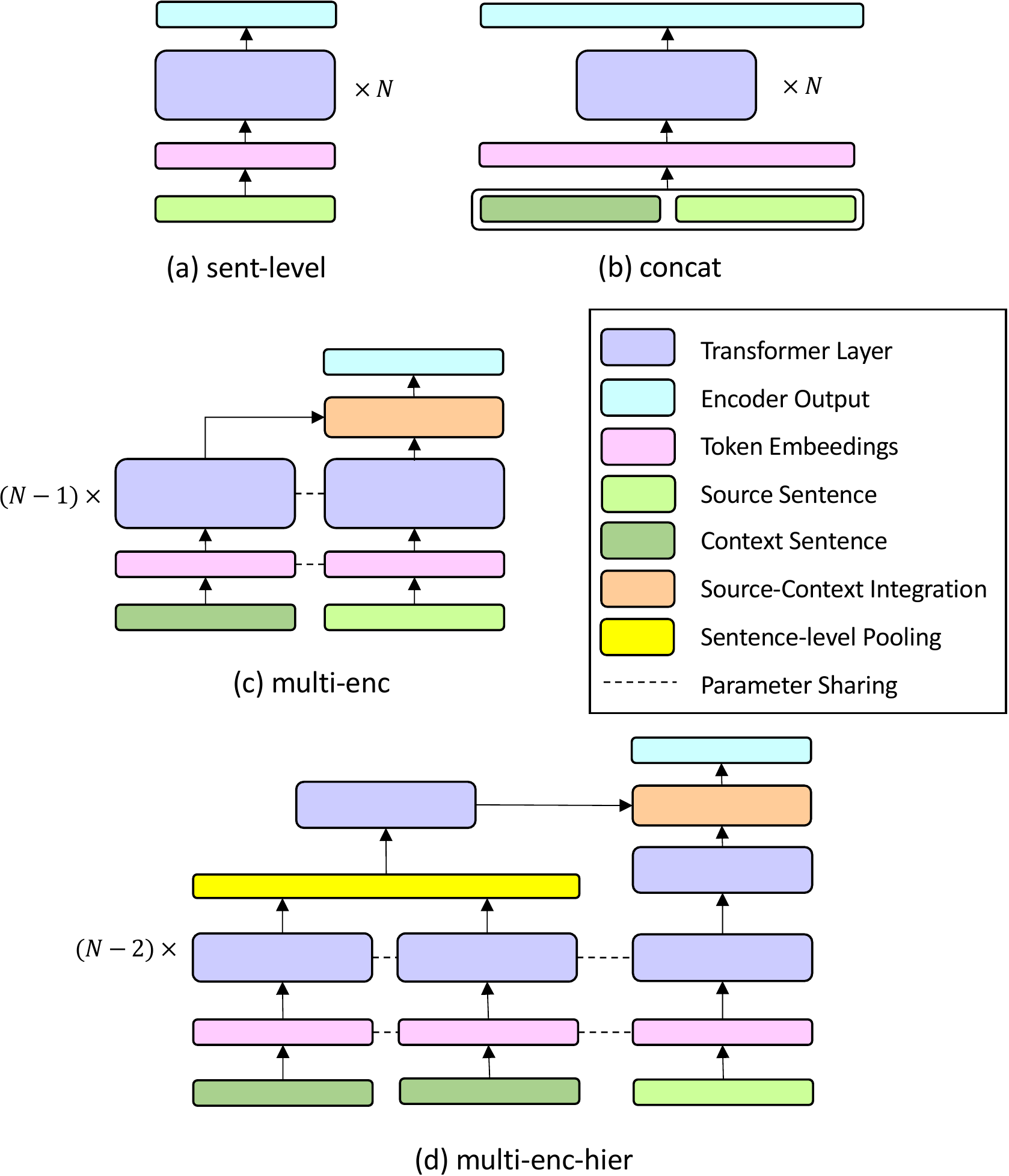}}
    \caption{The structure of compared context-aware NMT models.}
    \label{fig:models}
\end{figure}

Throughout this work, we consider Transformer \cite{Vaswani2017Attention} as a base model architecture by following the majority of the recent works on context-aware NMT. Transformer consists of a stack of self-attentional layers in which a self-attention module is followed by a feed-forward module for each layer. Here we list four Transformer-based configurations that we used in the experiments:

\begin{itemize}
    \item \textbf{sent-level}:
    As a baseline, we have experimented with the basic Transformer model which does not use any contextual sentences. 
    
    \item \textbf{concat}: 
    This is a straightforward approach to incorporate contextual sentences without modifying the Transformer model \cite{tiedemann-scherrer-2017-neural}. This concatenates all contextual sentences and an input sentence with special tokens between sentences.
    
    \item \textbf{multi-enc}:
    This has an extra encoder for encoding contextual sentences separately. We follow the model introduced in \cite{voita-etal-2018-context} which obtain a hidden representation of contextual sentences by weight-shared Transformer encoder. The model combines the encoded source and context representations using a source-to-context attention mechanism and a gated summation.

    \item \textbf{multi-enc-hier}: 
    To represent multiple contextual sentences effectively, hierarchical encoders for contextual sentences have been proposed \cite{miculicich-etal-2018-document,Yun-Hwang2020Improving}. In this configuration, the context representation is calculated in token-level first, then finally processed in sentence-level. We experimented with the model of \cite{Yun-Hwang2020Improving} in this paper.

\end{itemize}
All the model structures are described in Figure \ref{fig:models}.

\section{Our Method: CorefCL}
In this section, we explain main idea of CorefCL, a data augmentation and contrastive learning scheme leveraging coreference between the source and contextual sentences.

\subsection{Data Augmentation Using Coreference}
Generally, constrastive learning encourages a model to discriminate ground-truth and contrastive (negative) examples. In existing works, a number of approaches have been studied for obtaining contrastive examples:
\begin{itemize}
    \item Corrupting the sentence by randomly removing or replacing one or more tokens in the sentence. \cite{yang-etal-2019-reducing}
    \item Choosing irrelevant example in the batch or dataset.  \cite{pan-etal-2021-contrastive}
    \item Perturbations on representation space. Usually output vector of encoder or decoder is used. \cite{Lee2021contrastive}
\end{itemize}

\begin{figure*}[t]
\begin{center}
    \resizebox{0.65\textwidth}{!}{\includegraphics{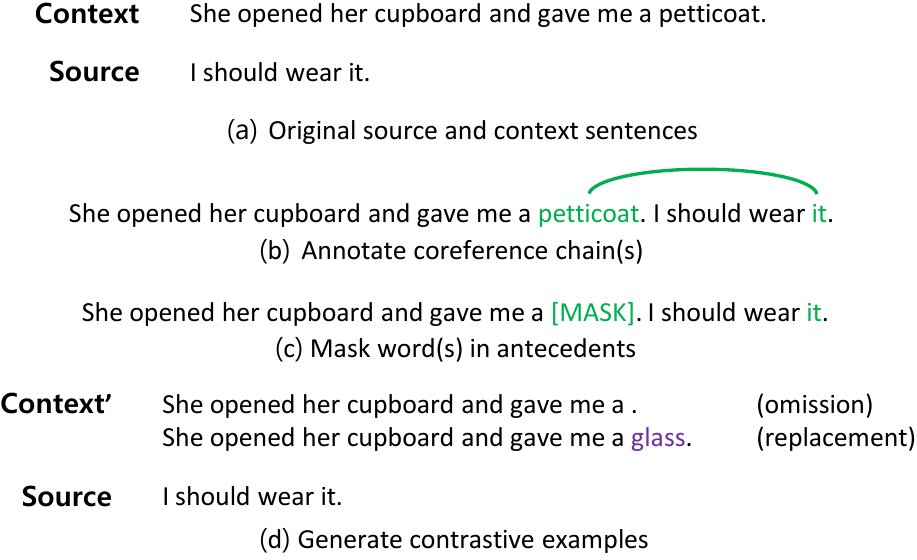}}
    \caption{Data augmentation process of CorefCL.}
    \label{fig:augmentation}
\end{center}
\end{figure*}

CorefCL basically takes a similar approach as the first one, by the sentence corruption. However, unlike previous works that modify the source sentence, CorefCL modifies the contextual sentences to form contrastive examples. Specifically, we corrupt cross-sentential coreference mentions which occur between the source and its contextual sentences. This is based on the intuition that coreference is one of the core components of coherent translation.

More formally, steps to forming contrastive examples in CorefCL are as follows (see also Figure \ref{fig:augmentation}):

\begin{enumerate}
  \item Annotate the source documents automatically. We use the NeuralCoref\footnote{https://github.com/huggingface/neuralcoref} to identify the coreference mentions between the source and its previous sentences as contextual sentences
  \item Filter the examples with cross-sentential coreference chain(s) between the source and contextual sentences. Around 20 to 30\% of the training corpus is annotated in this way. See Section \ref{exp:datasets} for details
  \item For each coreference chain, mask every word in the antecedents with a special token. We also keep the original examples for training
  \item Masked words are replaced randomly with other words in vocabulary (\textit{word replacement}), or omitted (\textit{word omission})
\end{enumerate}

In the experiments, we take both of the corruption strategies. Precisely, the masked words are removed with a probability of 0.5, or randomly replaced otherwise. We found that this method is more effective compared to the methods using only one of the two corruption strategies. Please refer to the ablation study in Section \ref{exp:analysis} for more details.

\subsection{Contrastive Learning for Context-aware NMT}
Context-aware NMT models can implicitly capture dependencies between the source and contextual sentences. CorefCL introduces a max-margin contrastive learning loss to train the model to explicitly discriminate inconsistent contexts. This contrastive loss also encourages a model to be more sensitive to the contents of contextual sentences.

Formally, given the source $\mathbf{x}$, target $\mathbf{y}$, $n$ contextual sentences $C = [\mathbf{c}_1, \cdots, \mathbf{c}_n]$ in the data $\mathcal{D}$, we first train the model by minimizing a negative log-likelihood loss, which is a common MT loss:

\[ \mathcal{L}_{MT} = \sum_{(\mathbf{x},\mathbf{y},C) \in \mathcal{D}}
- \mathrm{log} P ( \mathbf{y} | \mathbf{x}, C ).\]

Once the model is trained with MT loss, we fine-tune the model with a contrastive loss. With a contrastive version of context $\tilde{C}$, our contrastive learning objective is minimizing a max-margin loss \cite{huang-etal-2018-large, yang-etal-2019-reducing}:

\begin{multline*}
\mathcal{L}_{CL} = \sum_{(\mathbf{x},\mathbf{y},C,\tilde{C}) \in \mathcal{D}} 
\mathrm{max} \{ \eta + \mathrm{log} P ( \mathbf{y} | \mathbf{x}, \tilde{C} ) \\
- \mathrm{log} P ( \mathbf{y} | \mathbf{x}, C ), 0 \}.
\end{multline*}

Minimizing $\mathcal{L}_{CL}$ encourages the log-likelihood of the ground-truth to be at least $\eta$ larger than that of the contrastive examples. In our formulation, we want the model to be more sensitive to the subtle changes in the contextual sentences.

The contrastive loss is jointly optimized with MT loss since we empirically found that the joint optimization has yielded better performance than minimizing CL loss only as similar to \cite{Yu2020thedeepmind}:

\[ \mathcal{L} = (1-\alpha)\mathcal{L}_{MT} + \alpha \mathcal{L}_{CL}, \]

where $\alpha \in [0,1]$ is a weight for balancing between contrastive learning and MT loss.  For simplicity, we fixed $\alpha$  during fine-tuning.

\section{Experiments}
\subsection{Datasets}\label{exp:datasets}

\begin{table*}[ht]
\centering
\begin{tabular}{l|l|l|l|ll}
\hline
System & WMT & OpenSubtitles & IWSLT &  \multicolumn{2}{c}{En-Ko Subtitles}\\
& & & & detok. & char. \\
\hline
sent-level & 22.7 & 27.6 & 29.3 & 8.6 & 19.2 \\
\hline
concat & 22.4 & 28.3 & 29.7 & 9.3 & 22.1 \\
+ CorefCL & 23.5 (+1.1) & 29.1 (+0.8) & 30.9 (+1.3) & \underline{10.9 (+1.6)} & \underline{24.9 (+2.8)} \\
multi-enc & 23.1  & 28.6 & 29.8 & 9.2 & 21.7 \\
+ CorefCL & \underline{24.3 (+1.2)} & \underline{29.8 (+1.4)} & \underline{31.1 (+1.3)} & \underline{10.8 (+1.6)} & 24.4 (+2.7) \\
multi-enc-hier & 24.4 & 29.1 & 30.0 & 10.3 & 23.1 \\
+ CorefCL & 25.4 (+1.0) & 30.2 (+1.1) & 31.1 (+1.2) & 11.7 (+1.4) & 25.7 (+2.6) \\
\hline
\end{tabular}
\caption{\label{table:overall-bleu}
Corpus-level BLEU scores of compared models on different tasks. For the En-Ko subtitles task, we list both detokenized (detok.) and character-level (char.) scores. Improvements by CorefCL are denoted in (). Underlined score means that the model has the largest BLEU improvements among models in the same task.
}
\end{table*}

We experimented with CorefCL on various document-level parallel datasets: i) 3 English-German datasets including WMT document-level news translation\footnote{http://www.statmt.org/wmt19/translation-task.html} \cite{barrault-etal-2019-findings}, IWSLT TED talk \footnote{https://wit3.fbk.eu/home} \cite{Cettolo2017overview-iwslt}, OpenSubtitles'18\footnote{https://opus.nlpl.eu/OpenSubtitles-v2018.php}  \cite{lison-etal-2018-opensubtitles2018}, and ii) our web-crawled English-Korean subtitles corpus.

For all tasks, we take every 2 preceding sentences as contextual sentences and we only consider sentences only within the same document (article, talk, movie, one episode of TV programs, etc.) of the source sentence. If split of the validation and the test set is not presented in the data, we apply document-based split to ensure that training and validation/test data is well-separated. Details of datasets are listed as follows:

\textbf{WMT} We use a set of parallel corpora annotated with document boundaries which is released in WMT'19 news translation task. Specifically, we combine Europarl v9, News Commentary v14, and MODEL-RAPID to form a training set containing 3.7$M$ examples and 0.85$M$ with cross-sentential coreferences. For validation and test sets, we used newstest2013 and newstest2019 which contain 3.05$k$ and 2.14$k$ examples respectively.

\textbf{IWSLT} The IWSLT dataset consists of transcriptions of TED talks in a variety of languages. We used the 2017 version of the training set, a combination of dev2010, tst2010, tst2015 as a validation set, and tst2017 as a test set. The resulting dataset consists of 232$k$ (50.3$k$ with cross-sentential coreferences), 3.5$k$, 1.2$k$ examples of train, dev, test sets respectively. 

\textbf{OpenSubtitles} We also choose the English-German pair of OpenSubtitles2018  corpora. The raw corpus contains 24.4$M$ parallel sentences. We follow the filtering methods in \cite{voita-etal-2019-good} by removing pairs that have a time overlap of subtitle frames less than 0.9. We also use separate documents for validation / test sets, resulting 3.9$M$ (1.01$M$ with cross-sentential coreferences), 40.7$k$, 40.5$k$ examples for train / validation / test sets respectively.

\textbf{En-Ko Subtitles} For English-Korean experiments, we first crawled approximately 6.1$k$ bilingual subtitle files from websites such as GomLab.com. Since sentence pairs of these subtitles are already soft-aligned by the creators so we applied a simple time-code based heuristics to filter examples. The final data contains 1.6$M$ (0.24$M$ with cross-sentential coreferences), 155.6$k$, and 18.1$k$ examples of consecutive sentences in the training, validation, and test sets respectively. 

For preprocessing, all English and German corpus is tokenized first with Moses \cite{koehn-etal-2007-moses} tokenizer\footnote{https://github.com/moses-smt/mosesdecoder}. We then apply the BPE \cite{sennrich-etal-2016-neural} using SentencePiece\footnote{https://github.com/google/sentencepiece}, and the size of the merge operation is approximately 16.5$k$. We also put a special token [BOC] at the beginning of contextual sentences to differentiate them from the source sentences. 

\subsection{Settings}
We use model hyperparameters, such as the size of hidden dimensions and the number of hidden layers as same the \texttt{transformer-base} \cite{Vaswani2017Attention}, since all of the compared models are based on Transformer. Specifically, we set 512 as the hidden dimension, the number of layers is 6, the number of attention heads is 8, and the dropout rate is set to 0.1.

All models are trained with ADAM \cite{kingma2014adam} with different learning rates for each dataset. We employ early stopping of the training when the MT loss on the validation set does not improve. We start training each baseline model from scratch with random initialization and document-level dataset. Note that all the baseline models are not trained using iterative training as \cite{zhang-etal-2018-improving, huo-etal-2020-diving} which first trains the model from sentence-level task first, then document-level task.
All the evaluated models are implemented on top of the transformer\footnote{https://github.com/huggingface/transformers} framework.

We measure the translation quality by the BLEU score \cite{papineni-etal-2002-bleu}. For scoring BLEU, we use the sacreBLEU \cite{post-2018-call} case-sensitive, detokenized scores for En-De, and case-insensitive scores with \texttt{intl} tokenizer for En-Ko task. We also report case-insensitive char-level scores on En-Ko for comparison.

\subsection{Overall BLEU Evaluation}
We display the corpus-level test BLEU scores of all compared models on different tasks on Table \ref{table:overall-bleu}. Among the baseline systems, all context-aware models show moderate improvements over the sentence-level (sent-level) baseline. These results are comparable to that of \citet{huo-etal-2020-diving} on the IWSLT task except for multi-enc-hier, and \citet{Yun-Hwang2020Improving} on OpenSubtitles task. One exception is a single-encoder model (concat) on WMT task, which seems due to the longer average sentence length.

We evaluated CorefCL by fine-tuning the context-aware models. Results show that models with CorefCL outperformed their vanilla counterparts, with the BLEU gain of up to 1.4 in En-De tasks, and 1.6/2.8 (detokenized/char-level BLEU) in the En-Ko subtitles task. 

We observed that while CorefCL consistently improves BLEU on all tasks, it achieves better results on IWSLT and En-Ko subtitles tasks. Since improvements on much larger datasets like WMT and OpenSubtitles are smaller, we suggest that CorefCL also works as a regularization. 

\subsection{Results on English-German Contrastive Evaluation Set}

\begin{table}[h]
\centering
\begin{tabular}{l|cc|cc}
\hline
System & \multicolumn{4}{c}{Trained on} \\
 & \multicolumn{2}{c}{WMT} & \multicolumn{2}{c}{OpenSubtitles} \\
 & BLEU & Acc. & BLEU & Acc. \\
\hline
sent-level & 19.3 & 47.9 & 29.6 & 48.4 \\
\hline
concat & 19.9 & 49.7& 30.5 & 54.4 \\
+ CorefCL & 20.3 & 51.2 & 32.3 & 57.9\\
multi-enc-hier & 20.4 & 50.9& 31.7 & 57.3\\
+ CorefCL & 21.9 & 52.4 & 33.6 & 60.5\\
\hline
\end{tabular}
\caption{\label{table:contrastive_set}
BLEU and pronoun resolution accuracies on ContraPro \cite{muller-etal-2018-large} En-De contrastive test set.
}
\end{table}
To assess how CorefCL improves the ability to deal with pronoun-related translations more in detail, we experiment our method with ContraPro\footnote{https://github.com/ZurichNLP/ContraPro}. ContraPro is a contrastive test suit for En-De pronoun translation introduced by \citet{muller-etal-2018-large}. The evaluation is done by letting the model scores the German sentence with correct and incorrect pronoun translation, given the source and contextual English sentence. The accuracy is calculated by counting the number of correctly scored examples (i.e. correct examples that received a higher score than their incorrect counterpart).  

We evaluate the models trained with WMT and OpenSubtitles tasks. We also list BLEU scores of En-De translation using the English source text in ContraPro. As shown in Table. \ref{table:contrastive_set}, CorefCL significantly improves the baselines in scoring accuracy for all models by up to 5.5\%, as well as slight improvements in BLEU scores. 

One interesting finding is that CorefCL also achieved substantial accuracy gain on the models trained on WMT. Since the ContraPro is created from OpenSubtitles, WMT-trained models would yield lower performance because of domain shift between training and testing. Table.\ref{table:contrastive_set} clearly shows the performance drop in BLEU, nevertheless, moderate improvements in accuracy can also be observed on WMT-trained models.

\subsection{Analysis} \label{exp:analysis}
\begin{table}[h]
\centering
\begin{tabular}{l|c|c}
\hline
System & BLEU & Accuracy \\
\hline
multi-enc-hier & 31.7 & 57.3\\
+ CorefCL & 33.6 & 60.5\\
- Word omission & 32.4 & 59.4 \\
- Word replacement & 32.3 & 58.6 \\
\hline
\end{tabular}
\caption{\label{table:analysis}
Ablation study on coreference corruption strategy. All systems are trained on OpenSubtitle English-German dataset and evaluated on ContraPro.
}
\end{table}
\textbf{Ablation Study} CorefCL uses the two corruption strategies for generating contrastive coreference mentions; word omission and word replacement. To make a better understanding of influence of these strategies, we evaluate CorefCL of different settings of these strategies. 

As shown in Table.\ref{table:analysis}, using both types of corruptions results in better performance. Removing one of the two strategies slightly degrades both the pronoun resolution accuracy and BLEU. Although not being significant, removing the word replacement has more impact on accuracy. This suggests that a standard context-aware model, at least for multi-enc-hier is less sensitive to the word substitution. The word replacement strategy can complement this behavior as resulted in better performance.

\begin{figure}[h]
    \resizebox{0.95\columnwidth}{!}{\includegraphics{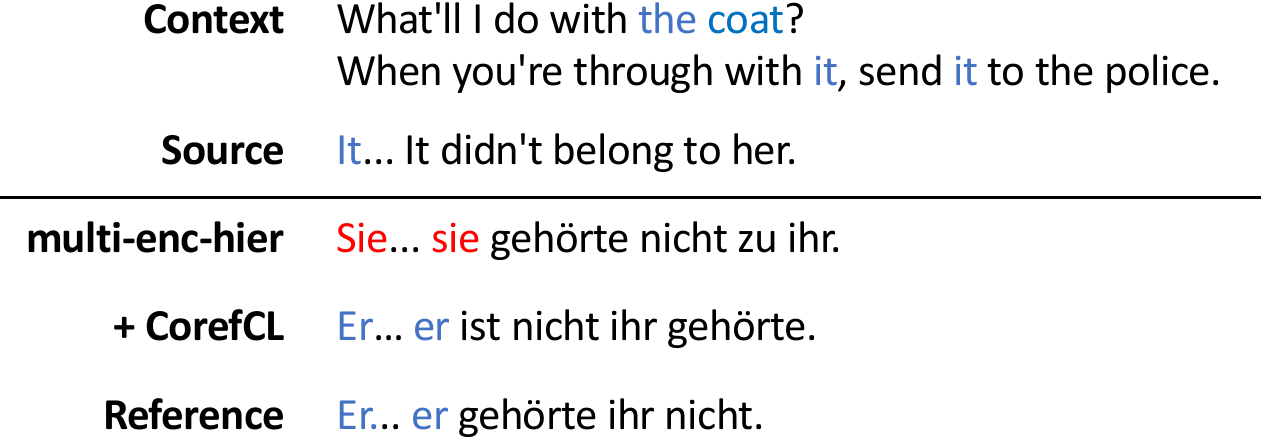}}
    \caption{Example translation with and without CorefCL. }
    \label{fig:sample}
\end{figure}
\textbf{Qualitative Example} We display a sample from ContraPro corpus and its translations made by multi-enc-hier model trained with OpenSubtitle task. In this example, since "coat" is translated as \textit{Mantel} which is a masculine noun thus \textit{Er} would be adequate translation of "It" instead of \textit{Sie} which is feminine. While multi-enc-hier incorrectly translated "It" as \textit{Sie}, the model fine-tuned with CorefCL correctly resolved it as \textit{Er}.

In practice, context-aware models do not leverage target-side contexts struggle to maintain these kinds of coreference consistency \cite{muller-etal-2018-large, lapshinova-koltunski-etal-2019-analysing} because of the asymmetric nature of grammatical components and data distributions. Results show that CorefCL can complement the limitation of source-only context-aware models.

\section{Conclusions and Future Work}
We have presented a data augmentation and contrastive learning scheme based on coreference for context-aware NMT. By leveraging coreference mentions between the source and target sentence, CorefCL effectively generates contrastive examples for applying contrastive learning on context-aware NMT models. In the experiments, CorefCL consistently improves the translation quality and pronoun resolution accuracy.

As future work, we plan to extend CorefCL to target contexts since maintaining coreference consistency needs both the source and the target contexts. It would be also interesting that applying CorefCL for fine-tuning pre-trained big language models like BART \cite{lewis-etal-2020-bart} or T5 \cite{raffel2020exploring} for downstream document-level MT tasks.

\bibliography{anthology,custom}

\begin{thebibliography}{44}
\expandafter\ifx\csname natexlab\endcsname\relax\def\natexlab#1{#1}\fi

\bibitem[{Barrault et~al.(2019)Barrault, Bojar, Costa-juss{\`a}, Federmann,
  Fishel, Graham, Haddow, Huck, Koehn, Malmasi, Monz, M{\"u}ller, Pal, Post,
  and Zampieri}]{barrault-etal-2019-findings}
Lo{\"\i}c Barrault, Ond{\v{r}}ej Bojar, Marta~R. Costa-juss{\`a}, Christian
  Federmann, Mark Fishel, Yvette Graham, Barry Haddow, Matthias Huck, Philipp
  Koehn, Shervin Malmasi, Christof Monz, Mathias M{\"u}ller, Santanu Pal, Matt
  Post, and Marcos Zampieri. 2019.
\newblock \href {https://doi.org/10.18653/v1/W19-5301} {Findings of the 2019
  conference on machine translation ({WMT}19)}.
\newblock In \emph{Proceedings of the Fourth Conference on Machine Translation
  (Volume 2: Shared Task Papers, Day 1)}, pages 1--61, Florence, Italy.
  Association for Computational Linguistics.

\bibitem[{Cettolo et~al.(2017)Cettolo, Federico, Bentivogli, Niehues,
  St{\"{u}}ker, Sudoh, Yoshino, and Federmann}]{Cettolo2017overview-iwslt}
Mauro Cettolo, Marcello Federico, Luisa Bentivogli, Jan Niehues, Sebastian
  St{\"{u}}ker, Katsuhito Sudoh, Koichiro Yoshino, and Christian Federmann.
  2017.
\newblock {Overview of the IWSLT 2017 Evaluation Campaign}.
\newblock In \emph{Proceedings of the 14th International Workshop on Spoken
  Language Translation (IWSLT)}, pages 1--14.

\bibitem[{Chen et~al.(2020)Chen, Kornblith, Norouzi, and
  Hinton}]{chen2020simple}
Ting Chen, Simon Kornblith, Mohammad Norouzi, and Geoffrey Hinton. 2020.
\newblock \href {http://proceedings.mlr.press/v119/chen20j.html} {A simple
  framework for contrastive learning of visual representations}.
\newblock In \emph{Proceedings of the 37th International Conference on Machine
  Learning}, volume 119 of \emph{Proceedings of Machine Learning Research},
  pages 1597--1607. PMLR.

\bibitem[{Cho et~al.(2014)Cho, van Merri{\"e}nboer, Gulcehre, Bahdanau,
  Bougares, Schwenk, and Bengio}]{cho-etal-2014-learning}
Kyunghyun Cho, Bart van Merri{\"e}nboer, Caglar Gulcehre, Dzmitry Bahdanau,
  Fethi Bougares, Holger Schwenk, and Yoshua Bengio. 2014.
\newblock \href {https://doi.org/10.3115/v1/D14-1179} {Learning phrase
  representations using {RNN} encoder{--}decoder for statistical machine
  translation}.
\newblock In \emph{Proceedings of the 2014 Conference on Empirical Methods in
  Natural Language Processing ({EMNLP})}, pages 1724--1734, Doha, Qatar.
  Association for Computational Linguistics.

\bibitem[{Guillou et~al.(2018)Guillou, Hardmeier, Lapshinova-Koltunski, and
  Lo{\'a}iciga}]{guillou-etal-2018-pronoun}
Liane Guillou, Christian Hardmeier, Ekaterina Lapshinova-Koltunski, and Sharid
  Lo{\'a}iciga. 2018.
\newblock \href {https://doi.org/10.18653/v1/W18-6435} {A pronoun test suite
  evaluation of the {E}nglish{--}{G}erman {MT} systems at {WMT} 2018}.
\newblock In \emph{Proceedings of the Third Conference on Machine Translation:
  Shared Task Papers}, pages 570--577, Belgium, Brussels. Association for
  Computational Linguistics.

\bibitem[{Huang et~al.(2018)Huang, Li, Ping, and Huang}]{huang-etal-2018-large}
Jiaji Huang, Yi~Li, Wei Ping, and Liang Huang. 2018.
\newblock \href {https://doi.org/10.18653/v1/D18-1150} {Large margin neural
  language model}.
\newblock In \emph{Proceedings of the 2018 Conference on Empirical Methods in
  Natural Language Processing}, pages 1183--1191, Brussels, Belgium.
  Association for Computational Linguistics.

\bibitem[{Huo et~al.(2020)Huo, Herold, Gao, Dahlmann, Khadivi, and
  Ney}]{huo-etal-2020-diving}
Jingjing Huo, Christian Herold, Yingbo Gao, Leonard Dahlmann, Shahram Khadivi,
  and Hermann Ney. 2020.
\newblock \href {https://aclanthology.org/2020.wmt-1.71} {Diving deep into
  context-aware neural machine translation}.
\newblock In \emph{Proceedings of the Fifth Conference on Machine Translation},
  pages 604--616, Online. Association for Computational Linguistics.

\bibitem[{Jean et~al.(2017)Jean, Lauly, Firat, and Cho}]{jean2017does}
Sebastien Jean, Stanislas Lauly, Orhan Firat, and Kyunghyun Cho. 2017.
\newblock Does neural machine translation benefit from larger context?
\newblock \emph{arXiv preprint arXiv:1704.05135}.

\bibitem[{Kim et~al.(2019)Kim, Tran, and Ney}]{kim-etal-2019-document}
Yunsu Kim, Duc~Thanh Tran, and Hermann Ney. 2019.
\newblock \href {https://doi.org/10.18653/v1/D19-6503} {When and why is
  document-level context useful in neural machine translation?}
\newblock In \emph{Proceedings of the Fourth Workshop on Discourse in Machine
  Translation (DiscoMT 2019)}, pages 24--34, Hong Kong, China. Association for
  Computational Linguistics.

\bibitem[{Kingma and Ba(2014)}]{kingma2014adam}
Diederik~P Kingma and Jimmy Ba. 2014.
\newblock \href {https://arxiv.org/abs/1412.6980} {Adam: A method for
  stochastic optimization}.
\newblock \emph{arXiv preprint}, arXiv:1412.6980.

\bibitem[{Koehn et~al.(2007)Koehn, Hoang, Birch, Callison-Burch, Federico,
  Bertoldi, Cowan, Shen, Moran, Zens, Dyer, Bojar, Constantin, and
  Herbst}]{koehn-etal-2007-moses}
Philipp Koehn, Hieu Hoang, Alexandra Birch, Chris Callison-Burch, Marcello
  Federico, Nicola Bertoldi, Brooke Cowan, Wade Shen, Christine Moran, Richard
  Zens, Chris Dyer, Ond{\v{r}}ej Bojar, Alexandra Constantin, and Evan Herbst.
  2007.
\newblock \href {https://aclanthology.org/P07-2045} {{M}oses: Open source
  toolkit for statistical machine translation}.
\newblock In \emph{Proceedings of the 45th Annual Meeting of the Association
  for Computational Linguistics Companion Volume Proceedings of the Demo and
  Poster Sessions}, pages 177--180, Prague, Czech Republic. Association for
  Computational Linguistics.

\bibitem[{Lample et~al.(2018)Lample, Ott, Conneau, Denoyer, and
  Ranzato}]{lample-etal-2018-phrase}
Guillaume Lample, Myle Ott, Alexis Conneau, Ludovic Denoyer, and Marc{'}Aurelio
  Ranzato. 2018.
\newblock \href {https://doi.org/10.18653/v1/D18-1549} {Phrase-based {\&}
  neural unsupervised machine translation}.
\newblock In \emph{Proceedings of the 2018 Conference on Empirical Methods in
  Natural Language Processing}, pages 5039--5049, Brussels, Belgium.
  Association for Computational Linguistics.

\bibitem[{Lapshinova-Koltunski et~al.(2019)Lapshinova-Koltunski,
  Espa{\~n}a-Bonet, and van
  Genabith}]{lapshinova-koltunski-etal-2019-analysing}
Ekaterina Lapshinova-Koltunski, Cristina Espa{\~n}a-Bonet, and Josef van
  Genabith. 2019.
\newblock \href {https://doi.org/10.18653/v1/D19-6501} {Analysing coreference
  in transformer outputs}.
\newblock In \emph{Proceedings of the Fourth Workshop on Discourse in Machine
  Translation (DiscoMT 2019)}, pages 1--12, Hong Kong, China. Association for
  Computational Linguistics.

\bibitem[{Lapshinova-Koltunski et~al.(2020)Lapshinova-Koltunski, Krielke, and
  Hardmeier}]{lapshinova-koltunski-etal-2020-coreference}
Ekaterina Lapshinova-Koltunski, Marie-Pauline Krielke, and Christian Hardmeier.
  2020.
\newblock \href {https://aclanthology.org/2020.crac-1.15} {Coreference
  strategies in {E}nglish-{G}erman translation}.
\newblock In \emph{Proceedings of the Third Workshop on Computational Models of
  Reference, Anaphora and Coreference}, pages 139--153, Barcelona, Spain
  (online). Association for Computational Linguistics.

\bibitem[{L{\"a}ubli et~al.(2018)L{\"a}ubli, Sennrich, and
  Volk}]{laubli-etal-2018-machine}
Samuel L{\"a}ubli, Rico Sennrich, and Martin Volk. 2018.
\newblock \href {https://doi.org/10.18653/v1/D18-1512} {Has machine translation
  achieved human parity? a case for document-level evaluation}.
\newblock In \emph{Proceedings of the 2018 Conference on Empirical Methods in
  Natural Language Processing}, pages 4791--4796, Brussels, Belgium.
  Association for Computational Linguistics.

\bibitem[{Lee et~al.(2021)Lee, Lee, and Hwang}]{Lee2021contrastive}
Seanie Lee, Dong~Bok Lee, and Sung~Ju Hwang. 2021.
\newblock \href {http://arxiv.org/abs/2012.07280} {{Contrastive Learning with
  Adversarial Perturbations for Conditional Text Generation}}.
\newblock In \emph{ICLR}, pages 1--25.

\bibitem[{Lewis et~al.(2020)Lewis, Liu, Goyal, Ghazvininejad, Mohamed, Levy,
  Stoyanov, and Zettlemoyer}]{lewis-etal-2020-bart}
Mike Lewis, Yinhan Liu, Naman Goyal, Marjan Ghazvininejad, Abdelrahman Mohamed,
  Omer Levy, Veselin Stoyanov, and Luke Zettlemoyer. 2020.
\newblock \href {https://doi.org/10.18653/v1/2020.acl-main.703} {{BART}:
  Denoising sequence-to-sequence pre-training for natural language generation,
  translation, and comprehension}.
\newblock In \emph{Proceedings of the 58th Annual Meeting of the Association
  for Computational Linguistics}, pages 7871--7880, Online. Association for
  Computational Linguistics.

\bibitem[{Li et~al.(2020)Li, Liu, Wang, Jiang, Xiao, Zhu, Liu, and
  Li}]{li-etal-2020-multi-encoder}
Bei Li, Hui Liu, Ziyang Wang, Yufan Jiang, Tong Xiao, Jingbo Zhu, Tongran Liu,
  and Changliang Li. 2020.
\newblock \href {https://doi.org/10.18653/v1/2020.acl-main.322} {Does
  multi-encoder help? a case study on context-aware neural machine
  translation}.
\newblock In \emph{Proceedings of the 58th Annual Meeting of the Association
  for Computational Linguistics}, pages 3512--3518, Online. Association for
  Computational Linguistics.

\bibitem[{Lison et~al.(2018)Lison, Tiedemann, and
  Kouylekov}]{lison-etal-2018-opensubtitles2018}
Pierre Lison, J{\"o}rg Tiedemann, and Milen Kouylekov. 2018.
\newblock \href {https://aclanthology.org/L18-1275} {{O}pen{S}ubtitles2018:
  Statistical rescoring of sentence alignments in large, noisy parallel
  corpora}.
\newblock In \emph{Proceedings of the Eleventh International Conference on
  Language Resources and Evaluation ({LREC} 2018)}, Miyazaki, Japan. European
  Language Resources Association (ELRA).

\bibitem[{Miculicich et~al.(2018)Miculicich, Ram, Pappas, and
  Henderson}]{miculicich-etal-2018-document}
Lesly Miculicich, Dhananjay Ram, Nikolaos Pappas, and James Henderson. 2018.
\newblock \href {https://doi.org/10.18653/v1/D18-1325} {Document-level neural
  machine translation with hierarchical attention networks}.
\newblock In \emph{Proceedings of the 2018 Conference on Empirical Methods in
  Natural Language Processing}, pages 2947--2954, Brussels, Belgium.
  Association for Computational Linguistics.

\bibitem[{Mikolov et~al.(2013)Mikolov, Sutskever, Chen, Corrado, and
  Dean}]{mikolov2013distributed}
Tomas Mikolov, Ilya Sutskever, Kai Chen, Greg~S Corrado, and Jeff Dean. 2013.
\newblock \href
  {https://proceedings.neurips.cc/paper/2013/file/9aa42b31882ec039965f3c4923ce901b-Paper.pdf}
  {Distributed representations of words and phrases and their
  compositionality}.
\newblock In \emph{Advances in Neural Information Processing Systems},
  volume~26. Curran Associates, Inc.

\bibitem[{M{\"u}ller et~al.(2018)M{\"u}ller, Rios, Voita, and
  Sennrich}]{muller-etal-2018-large}
Mathias M{\"u}ller, Annette Rios, Elena Voita, and Rico Sennrich. 2018.
\newblock \href {https://doi.org/10.18653/v1/W18-6307} {A large-scale test set
  for the evaluation of context-aware pronoun translation in neural machine
  translation}.
\newblock In \emph{Proceedings of the Third Conference on Machine Translation:
  Research Papers}, pages 61--72, Brussels, Belgium. Association for
  Computational Linguistics.

\bibitem[{Ohtani et~al.(2019)Ohtani, Kamigaito, Nagata, and
  Okumura}]{ohtani-etal-2019-context}
Takumi Ohtani, Hidetaka Kamigaito, Masaaki Nagata, and Manabu Okumura. 2019.
\newblock \href {https://doi.org/10.18653/v1/D19-6505} {Context-aware neural
  machine translation with coreference information}.
\newblock In \emph{Proceedings of the Fourth Workshop on Discourse in Machine
  Translation (DiscoMT 2019)}, pages 45--50, Hong Kong, China. Association for
  Computational Linguistics.

\bibitem[{Pan et~al.(2021)Pan, Wang, Wu, and Li}]{pan-etal-2021-contrastive}
Xiao Pan, Mingxuan Wang, Liwei Wu, and Lei Li. 2021.
\newblock \href {https://doi.org/10.18653/v1/2021.acl-long.21} {Contrastive
  learning for many-to-many multilingual neural machine translation}.
\newblock In \emph{Proceedings of the 59th Annual Meeting of the Association
  for Computational Linguistics and the 11th International Joint Conference on
  Natural Language Processing (Volume 1: Long Papers)}, pages 244--258, Online.
  Association for Computational Linguistics.

\bibitem[{Papineni et~al.(2002)Papineni, Roukos, Ward, and
  Zhu}]{papineni-etal-2002-bleu}
Kishore Papineni, Salim Roukos, Todd Ward, and Wei-Jing Zhu. 2002.
\newblock \href {https://doi.org/10.3115/1073083.1073135} {{B}leu: a method for
  automatic evaluation of machine translation}.
\newblock In \emph{Proceedings of the 40th Annual Meeting of the Association
  for Computational Linguistics}, pages 311--318, Philadelphia, Pennsylvania,
  USA. Association for Computational Linguistics.

\bibitem[{Post(2018)}]{post-2018-call}
Matt Post. 2018.
\newblock \href {https://doi.org/10.18653/v1/W18-6319} {A call for clarity in
  reporting {BLEU} scores}.
\newblock In \emph{Proceedings of the Third Conference on Machine Translation:
  Research Papers}, pages 186--191, Brussels, Belgium. Association for
  Computational Linguistics.

\bibitem[{Raffel et~al.(2020)Raffel, Shazeer, Roberts, Lee, Narang, Matena,
  Zhou, Li, and Liu}]{raffel2020exploring}
Colin Raffel, Noam Shazeer, Adam Roberts, Katherine Lee, Sharan Narang, Michael
  Matena, Yanqi Zhou, Wei Li, and Peter~J. Liu. 2020.
\newblock \href {http://jmlr.org/papers/v21/20-074.html} {Exploring the limits
  of transfer learning with a unified text-to-text transformer}.
\newblock \emph{Journal of Machine Learning Research}, 21(140):1--67.

\bibitem[{Sennrich et~al.(2016{\natexlab{a}})Sennrich, Haddow, and
  Birch}]{sennrich-etal-2016-improving}
Rico Sennrich, Barry Haddow, and Alexandra Birch. 2016{\natexlab{a}}.
\newblock \href {https://doi.org/10.18653/v1/P16-1009} {Improving neural
  machine translation models with monolingual data}.
\newblock In \emph{Proceedings of the 54th Annual Meeting of the Association
  for Computational Linguistics (Volume 1: Long Papers)}, pages 86--96, Berlin,
  Germany. Association for Computational Linguistics.

\bibitem[{Sennrich et~al.(2016{\natexlab{b}})Sennrich, Haddow, and
  Birch}]{sennrich-etal-2016-neural}
Rico Sennrich, Barry Haddow, and Alexandra Birch. 2016{\natexlab{b}}.
\newblock \href {https://doi.org/10.18653/v1/P16-1162} {Neural machine
  translation of rare words with subword units}.
\newblock In \emph{Proceedings of the 54th Annual Meeting of the Association
  for Computational Linguistics (Volume 1: Long Papers)}, pages 1715--1725,
  Berlin, Germany. Association for Computational Linguistics.

\bibitem[{Smith(2017)}]{smith2017integrating}
Karin~Sim Smith. 2017.
\newblock On integrating discourse in machine translation.
\newblock In \emph{Proceedings of the Third Workshop on Discourse in Machine
  Translation}, pages 110--121.

\bibitem[{Stojanovski and Fraser(2018)}]{stojanovski-fraser-2018-coreference}
Dario Stojanovski and Alexander Fraser. 2018.
\newblock \href {https://doi.org/10.18653/v1/W18-6306} {Coreference and
  coherence in neural machine translation: A study using oracle experiments}.
\newblock In \emph{Proceedings of the Third Conference on Machine Translation:
  Research Papers}, pages 49--60, Brussels, Belgium. Association for
  Computational Linguistics.

\bibitem[{Sugiyama and Yoshinaga(2019)}]{sugiyama-yoshinaga-2019-data}
Amane Sugiyama and Naoki Yoshinaga. 2019.
\newblock \href {https://doi.org/10.18653/v1/D19-6504} {Data augmentation using
  back-translation for context-aware neural machine translation}.
\newblock In \emph{Proceedings of the Fourth Workshop on Discourse in Machine
  Translation (DiscoMT 2019)}, pages 35--44, Hong Kong, China. Association for
  Computational Linguistics.

\bibitem[{Sutskever et~al.(2014)Sutskever, Vinyals, and
  Le}]{Sutskever2014Sequence}
Ilya Sutskever, Oriol Vinyals, and Quoc~V. Le. 2014.
\newblock \href {http://arxiv.org/abs/1409.3215} {{Sequence to sequence
  learning with neural networks}}.
\newblock In \emph{Advances in Neural Information Processing Systems},
  December, pages 3104--3112, Montr{\'{e}}al, Canada.

\bibitem[{Tang et~al.(2018)Tang, M{\"u}ller, Rios, and
  Sennrich}]{tang-etal-2018-self}
Gongbo Tang, Mathias M{\"u}ller, Annette Rios, and Rico Sennrich. 2018.
\newblock \href {https://doi.org/10.18653/v1/D18-1458} {Why self-attention? a
  targeted evaluation of neural machine translation architectures}.
\newblock In \emph{Proceedings of the 2018 Conference on Empirical Methods in
  Natural Language Processing}, pages 4263--4272, Brussels, Belgium.
  Association for Computational Linguistics.

\bibitem[{Tiedemann and Scherrer(2017)}]{tiedemann-scherrer-2017-neural}
J{\"o}rg Tiedemann and Yves Scherrer. 2017.
\newblock \href {https://doi.org/10.18653/v1/W17-4811} {Neural machine
  translation with extended context}.
\newblock In \emph{Proceedings of the Third Workshop on Discourse in Machine
  Translation}, pages 82--92, Copenhagen, Denmark. Association for
  Computational Linguistics.

\bibitem[{Vaswani et~al.(2017)Vaswani, Shazeer, Parmar, Uszkoreit, Jones,
  Gomez, Kaiser, and Polosukhin}]{Vaswani2017Attention}
Ashish Vaswani, Noam Shazeer, Niki Parmar, Jakob Uszkoreit, Llion Jones,
  Aidan~N. Gomez, Lukasz Kaiser, and Illia Polosukhin. 2017.
\newblock \href {https://doi.org/10.1017/S0140525X16001837} {{Attention Is All
  You Need}}.
\newblock In \emph{Advances in Neural Information Processing Systems}.

\bibitem[{Voita et~al.(2019)Voita, Sennrich, and Titov}]{voita-etal-2019-good}
Elena Voita, Rico Sennrich, and Ivan Titov. 2019.
\newblock \href {https://doi.org/10.18653/v1/P19-1116} {When a good translation
  is wrong in context: Context-aware machine translation improves on deixis,
  ellipsis, and lexical cohesion}.
\newblock In \emph{Proceedings of the 57th Annual Meeting of the Association
  for Computational Linguistics}, pages 1198--1212, Florence, Italy.
  Association for Computational Linguistics.

\bibitem[{Voita et~al.(2018)Voita, Serdyukov, Sennrich, and
  Titov}]{voita-etal-2018-context}
Elena Voita, Pavel Serdyukov, Rico Sennrich, and Ivan Titov. 2018.
\newblock \href {https://doi.org/10.18653/v1/P18-1117} {Context-aware neural
  machine translation learns anaphora resolution}.
\newblock In \emph{Proceedings of the 56th Annual Meeting of the Association
  for Computational Linguistics (Volume 1: Long Papers)}, pages 1264--1274,
  Melbourne, Australia. Association for Computational Linguistics.

\bibitem[{Wu et~al.(2020)Wu, Wang, Gu, Khabsa, Sun, and Ma}]{wu2020clear}
Zhuofeng Wu, Sinong Wang, Jiatao Gu, Madian Khabsa, Fei Sun, and Hao Ma. 2020.
\newblock \href {http://arxiv.org/abs/2012.15466} {{CLEAR: Contrastive Learning
  for Sentence Representation}}.
\newblock \emph{arXiv preprint}, arXiv:2012.

\bibitem[{Yang et~al.(2019)Yang, Cheng, Liu, and Sun}]{yang-etal-2019-reducing}
Zonghan Yang, Yong Cheng, Yang Liu, and Maosong Sun. 2019.
\newblock \href {https://doi.org/10.18653/v1/P19-1623} {Reducing word omission
  errors in neural machine translation: A contrastive learning approach}.
\newblock In \emph{Proceedings of the 57th Annual Meeting of the Association
  for Computational Linguistics}, pages 6191--6196, Florence, Italy.
  Association for Computational Linguistics.

\bibitem[{Yu et~al.(2020)Yu, Sartran, Huang, Stokowiec, Donato, Srinivasan,
  Andreev, Ling, Mokra, {Dal Lago}, Doron, Young, Blunsom, and
  Dyer}]{Yu2020thedeepmind}
Lei Yu, Laurent Sartran, Po-Sen Huang, Wojciech Stokowiec, Domenic Donato,
  Srivatsan Srinivasan, Alek Andreev, Wang Ling, Sona Mokra, Agustin {Dal
  Lago}, Yotam Doron, Susannah Young, Phil Blunsom, and Chris Dyer. 2020.
\newblock \href {https://www.aclweb.org/anthology/2020.wmt-1.36} {{The DeepMind
  Chinese–English Document Translation System at WMT2020}}.
\newblock In \emph{Proceedings of the Fifth Conference on Machine Translation},
  pages 326--337, Online. Association for Computational Linguistics.

\bibitem[{Yun et~al.(2020)Yun, Hwang, and Jung}]{Yun-Hwang2020Improving}
Hyeongu Yun, Yongkeun Hwang, and Kyomin Jung. 2020.
\newblock \href {https://doi.org/10.1609/aaai.v34i05.6494} {Improving
  context-aware neural machine translation using self-attentive sentence
  embedding}.
\newblock \emph{Proceedings of the AAAI Conference on Artificial Intelligence},
  34(05):9498--9506.

\bibitem[{Zhang et~al.(2018)Zhang, Luan, Sun, Zhai, Xu, Zhang, and
  Liu}]{zhang-etal-2018-improving}
Jiacheng Zhang, Huanbo Luan, Maosong Sun, Feifei Zhai, Jingfang Xu, Min Zhang,
  and Yang Liu. 2018.
\newblock \href {https://doi.org/10.18653/v1/D18-1049} {Improving the
  transformer translation model with document-level context}.
\newblock In \emph{Proceedings of the 2018 Conference on Empirical Methods in
  Natural Language Processing}, pages 533--542, Brussels, Belgium. Association
  for Computational Linguistics.

\bibitem[{Zinsmeister et~al.(2017)Zinsmeister, Dipper, and
  Seiss}]{zinsmeister2017abstract}
Heike Zinsmeister, Stefanie Dipper, and Melanie Seiss. 2017.
\newblock \href {https://doi.org/10.5281/zenodo.1019699} {{Abstract pronominal
  anaphors and label nouns in German and English: Selected case studies and
  quantitative investigations}}.

\end{thebibliography}
\bibliographystyle{acl_natbib}

\end{document}